\documentclass[10pt,twocolumn,letterpaper]{article}

\usepackage[pagenumbers]{cvpr} 
\makeatletter
\@namedef{ver@everyshi.sty}{}
\makeatother
\usepackage{graphicx}
\usepackage{amsmath}
\usepackage{amssymb}
\usepackage{booktabs}
\usepackage{algorithm}
\usepackage[noend]{algpseudocode}
\usepackage[dvipsnames]{xcolor}
\usepackage{multirow}
\usepackage{dblfloatfix}
\usepackage{tikz}
\usetikzlibrary{tikzmark}
\usepackage[accsupp]{axessibility}

\usepackage[pagebackref,breaklinks,colorlinks]{hyperref}

\usepackage[capitalize]{cleveref}
\crefname{section}{Sec.}{Secs.}
\Crefname{section}{Section}{Sections}
\Crefname{table}{Table}{Tables}
\crefname{table}{Tab.}{Tabs.}

\def\cvprPaperID{7076} 
\def\confName{CVPR}
\def\confYear{2023}

\begin{document}

\title{Selective Structured State-Spaces for Long-Form Video Understanding}
\author{Jue Wang\quad
Wentao Zhu\quad
Pichao Wang\quad
Xiang Yu\quad
Linda Liu\quad
Mohamed Omar\quad
Raffay Hamid\\
Amazon Prime Video\\
{\tt\small $\{$juewangn, zhuwent, wpichao, xiangnyu, lindliu, omarmk, raffay$\}$@amazon.com}
}
\maketitle

\begin{abstract}

\noindent Effective modeling of complex spatiotemporal dependencies in long-form videos remains an open problem. The recently proposed Structured State-Space Sequence (S4) model with its linear complexity offers a promising direction in this space. However, we demonstrate that treating all image-tokens equally as done by S4 model can adversely affect its efficiency and accuracy. To address this limitation, we present a novel Selective S4 (\textit{i.e.}, S5) model that employs a lightweight mask generator to adaptively select informative image tokens resulting in more efficient and accurate modeling of long-term spatiotemporal dependencies in videos. Unlike previous mask-based token reduction methods used in transformers, our S5 model avoids the dense self-attention calculation by making use of the guidance of the momentum-updated S4 model. This enables our model to efficiently discard less informative tokens and adapt to various long-form video understanding tasks more effectively. However, as is the case for most token reduction methods, the informative image tokens could be dropped incorrectly. To improve the robustness and the temporal horizon of our model, we propose a novel long-short masked contrastive learning (LSMCL) approach that enables our model to predict longer temporal context using shorter input videos. We present extensive comparative results using three challenging long-form video understanding datasets (LVU, COIN and Breakfast), demonstrating that our approach consistently outperforms the previous state-of-the-art S4 model by up to $9.6\%$ accuracy while reducing its memory footprint by $23\%$.

\end{abstract}
\section{Introduction}
\begin{figure}[!htb]
    \centering
    \includegraphics[width=1.0\linewidth]{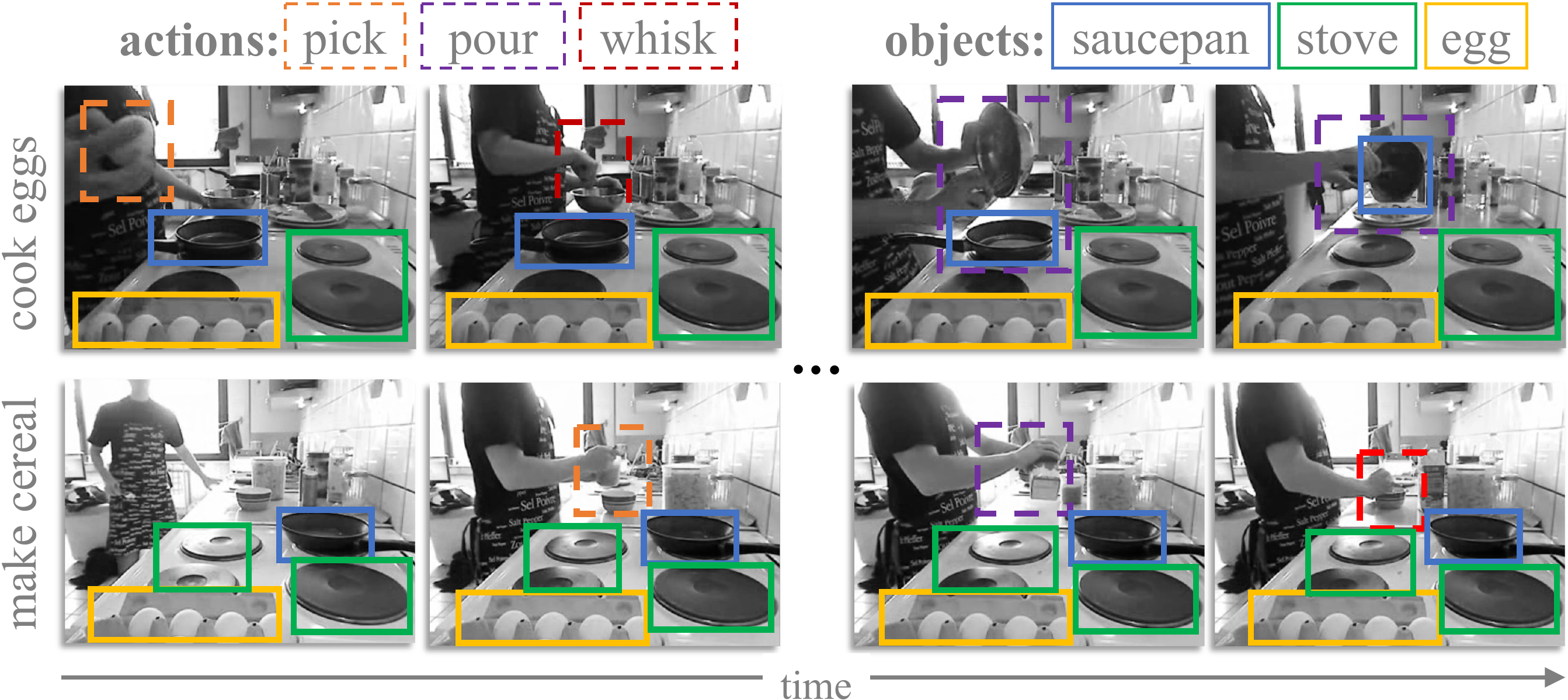}
    \vspace{-6mm}
    \caption{\textbf{Illustration of long-form videos --} Evenly sampled frames from two long-form videos, that have long duration (more than 1 minute) and distinct categories in the Breakfast~\cite{kuehne2014language} dataset (grayscale frames are shown for better visualization). The video on top shows the activity of making scrambled eggs, while the one on the bottom shows the activity of making cereal. These two videos heavily overlap in terms of objects (\textit{e.g.}, eggs, saucepan and stove), and actions (\textit{e.g.}, picking, whisking and pouring). To effectively distinguish these two videos, it is important to model long-term spatiotemporal dependencies, which is also the key in long-form video understanding.} 
    \label{Fig:example}
    \vspace{-2mm}
\end{figure}

\noindent Video understanding is an active research area where a variety of different models have been explored including \textit{e.g.}, two-stream networks~\cite{simonyan2014two,feichtenhofer2017temporal,feichtenhofer2017spatiotemporal}, recurrent neural networks~\cite{veeriah2015differential,zhang2017view,baccouche2011sequential} and $3$-D convolutional networks~\cite{tran2015learning,tran2018closer, tran2019video}.
%
However, most of these methods have primarily focused on short-form videos that are typically with a few seconds in length, and are not designed to model the complex long-term spatiotemporal dependencies often found in long-form videos (see Figure~\ref{Fig:example} for an illustrative example).
%
The recent vision transformer (ViT)~\cite{dosovitskiy2020image} has shown promising capability in modeling long-range dependencies,  and several variants~\cite{fan2021multiscale,bertasius2021space, arnab2021vivit, neimark2021video, patrick2021keeping, wang2022deformable, liu2021video} have successfully adopted the transformer architecture for video modeling. However, for a video with $\textrm{T}$ frames and $\textrm{S}$ spatial tokens, the complexity of standard video transformer architecture is $\mathcal{O}(\textrm{S}^2\textrm{T}^2)$, which poses prohibitively high computation and memory costs  when modeling long-form videos. Various attempts~\cite{wu2022memvit,sunlong2022} have been proposed to improve this efficiency, 
but the ViT pyramid architecture prevents them from developing long-term dependencies on low-level features.

In addition to ViT, a recent ViS4mer~\cite{islam2022long} method has tried to apply the Structured State-Spaces Sequence (S4) model~\cite{gu2021efficiently} as an effective way to model the long-term video dependencies.
%
%
%
%
However, by introducing simple masking techniques we empirically reveal that the S4 model can have different temporal reasoning preferences for different downstream tasks. This makes applying the same image token selection method as done by ViS4mer~\cite{islam2022long} for all long-form video understanding tasks suboptimal.



To address this challenge, we propose a cost-efficient adaptive token selection module, termed S$5$ (\textit{i.e.}, selective S$4$) model, which adaptively selects informative image tokens for the S4 model, thereby learning discriminative long-form video representations. 
%
Previous token reduction methods for efficient image transformers~\cite{yin2021adavit,yin2022vit, meng2022adavit,rao2021dynamicvit,liang2022not,wang2021efficient} heavily rely on a dense self-attention calculation, which makes them less effective in practice despite their theoretical guarantees about efficiency gains.
In contrast, our S5 model avoids the dense self-attention calculation by leveraging S4 features in a gumble-softmax sampling~\cite{jang2016categorical} based mask generator to adaptively select more informative image tokens. 
Our mask generator leverages S4 feature for its global sequence-context information and is further guided by the momentum distillation from the S4 model.

%
%

%

%
To further improve the robustness and the temporal predictability of our S5 model, we introduce a novel long-short mask contrastive learning (LSMCL) to pre-train our model. 
In LSMCL, randomly selected image tokens from long and short clips include the scenario that the less informative image tokens are chosen, and the representation of them are learned to match each other. As a result, the LSMCL not only significantly boosts the efficiency compared to the previous video contrastive learning methods~\cite{wang2022long,recasens2021broaden,feichtenhofer2021large}, but also increases the robustness of our S5 model when dealing with the mis-predicted image tokens. 
We empirically demonstrate that the S5 model with LSMCL pre-training can employ shorter-length clips to achieve on-par performance with using longer-range clips without incorporating LSMCL pre-training.

\vspace{0.1cm} \noindent We summarize our \textbf{key contributions} as the following: 


\noindent $\bullet$ We propose a Selective S4 (S5) model that leverages the global sequence-context information from S4 features to adaptively choose informative image tokens in a task-specific way.

\noindent $\bullet$ We introduce a novel long-short masked contrastive learning approach (LSMCL) that enables our model to be tolerant to the mis-predicted tokens and exploit longer duration spatiotemporal context by using shorter duration input videos, leading to improved robustness in the S5 model.


\noindent $\bullet$  We demonstrate that two proposed novel techniques (S5 model and LSMCL) are seamlessly suitable and effective for long-form video understanding, achieving the state-of-the-art performance on three challenging benchmarks. Notably, our method achieves up to $\textbf{9.6\%}$ improvement on LVU dataset compared to the previous state-of-the-art S4 method, while reducing the memory footprint by $\textbf{23\%}$.


\section{Related Work}
\noindent We discuss the literature with respect to the three most relevant fields: video understanding with long-form format, efficient token selection for vision transformer training, and self-supervised learning with videos.

\vspace{+2mm}
\noindent\textbf{a. Long-Form Video Modeling:} 
Transformers have shown excellent performance in modeling long-term dependencies,~\textit{e.g.}, in natural language processing (NLP)~\cite{brown2020language,dai2019transformer,devlin2018bert}. But the high computational cost caused by dense self-attention calculation becomes a bottleneck to apply in not only NLP but also computer vision. Much subsequent work~\cite{wang2022deformable, liu2021swin,choromanski2020rethinking,katharopoulos2020transformers,liu2021video,kitaev2020reformer,pan2021scalable} focuses on improving the transformer efficiency. However, they are not designed for dealing with plethora of spatial and temporal image tokens that are common in long-form video scenarios. LF-VILA~\cite{sunlong2022} develops a hierarchical feeding architecture to include more frames in the model, thus capturing longer temporal information. Similarly, MeMViT~\cite{wu2022memvit} better utilizes temporal information by emerging the previously cached ``memory" from the past. The pyramid structure leveraged by LF-VILA and MeMViT shows efficiency improvements, but may lose low-level spatial-temporal contextual information. 
Gu et al.~\cite{gu2021efficiently} proposed a Structured State-Space Sequence (S4) model, a novel alternative to CNNs or transformers, to model the long-range dependencies by simulating a linear time invariant (LTI) system. Subsequently, S4ND~\cite{nguyen2022s4nd} and ViS4mer~\cite{islam2022long} extend S4 model to the video classification task. ViS4mer~\cite{islam2022long} stacks multiple S4 layers with different scales in modeling long-form videos, and S4ND~\cite{nguyen2022s4nd} substitutes the traditional convolutional layer with the proposed S4ND layer in image and short-form video classification tasks. The equal importance assumption to all the image tokens by ViS4mer and S4ND can be further improved by introducing suitable token selection mechanisms, especially when dealing with the long-form input sequences. Consequently, we propose a token Selection S4 (S5) model to further enhance the efficiency while maintaining the long-form representation power.

\vspace{+2mm}
\noindent\textbf{b. Adaptive Token Selection:} 
Adaptive token selection is widely used to improve model efficiency. Traditional CNN methods such as SCsampler~\cite{korbar2019scsampler} filter informative clips by using motion and audio embeddings. Adaframe~\cite{wu2019adaframe} utilizes memory-augmented LSTMs as agents, which predict where to look in the next time step. AR-NET~\cite{meng2020ar} uses LSTM as decision maker to select useful frames and their resolutions. ~\cite{yin2021adavit,wang2021efficient,meng2022adavit,rao2021dynamicvit,liang2022not} apply this selection idea to transformers to adaptively select tokens for increased efficiency. For instance, STTS~\cite{wang2021efficient} leverages a token selection module, the named scorer network, to provide the importance score for each token and select the top-K frames with the highest scores. AdaViT~\cite{meng2022adavit} extends this idea to develop instance-specific policies, guiding the activation of patches, self-attention heads and transformer blocks. All of the above methods demonstrate how a light-weight token selection module can improve inference efficiency. However, these methods are essentially designed for images, and may require non-trivial adaptation to the long-form video scenarios,~\textit{i.e.}, the video-level long-range reasoning and computationally expensive self-attention calculation. To avoid this dense self-attention calculation, our proposed S5 model leverages S4 features to model the long-term dependencies and adaptively pick informative tokens.

\vspace{+2mm}
\noindent\textbf{c. Video Self-Supervised Learning (SSL):} Previous work on token reduction rarely considers the negative impact of mis-dropped tokens. EViT~\cite{liang2022not} simply fuses the  unattended tokens and concatenates with the remaining ones. From the recent successful image SSL works~\cite{he2020momentum,chen2020simple,grill2020bootstrap,he2022masked,chen2020exploring}, many follow-up works~\cite{tong2022videomae, wang2022long,recasens2021broaden,Feichtenhofer_large,feichtenhofer2022masked} learn discriminative video features with great generalization ability in downstream tasks. Specifically, LSTCL~\cite{wang2022long} and BraVe~\cite{recasens2021broaden} utilize long and short clips in the concept of SSL, which enables the model to learn an effective representation by predicting temporal context captured from a longer temporal extent. This essentially broadens the temporal horizon of the model for predicting longer temporal context with fewer from shorter input frames. In this paper, we adopt this idea with an additional random masking strategy to increase the efficiency of contrastive learning in long-form videos, and to further improve the robustness and the temporal predictability of our S5 model in downstream tasks. 
\section{Approach}
\noindent We start by summarizing Structured State-Space Sequence (S4)~\cite{gu2021efficiently} model and ViS4mer~\cite{islam2022long} ($\S$~\ref{ss:preliminaries}), followed by empirical analysis of S4 model in various long-form video understanding tasks ($\S$~\ref{ss:vis4mer_limitations}), and then providing the details of our proposed approach to address these limitations ($\S$~\ref{ss:adaptive_tokens} and $\S$~\ref{ss:LSMCL}).

\subsection{Preliminaries}
\label{ss:preliminaries}
\subsubsection{S4 Model}
\label{sss:preliminaries_s4}
\noindent Recall that a simple State-Space Model \textit{i.e.}, a linear time invariant (LTI) system can be written as:
\begin{align}
\begin{split}
\label{SSM}
\mathbf{x}^\prime (t) &= \mathbf{A}\mathbf{x}(t)+\mathbf{B}\mathbf{u}(t) \\
\mathbf{y}(t) &= \mathbf{C}\mathbf{x}(t) + \mathbf{D}\mathbf{u}(t).
\end{split}
\end{align}
Under deep learning setting, $\mathbf{A}$, $\mathbf{B}$ and $\mathbf{C}$ are learned via gradient descent while $+\mathbf{D}\mathbf{u}(t)$ is replaced by a residual connection. This formulation projects an input signal $\mathbf{u}(t)$ from one-dimensional space to an $\textrm{N}$-dimensional latent space $\mathbf{x}(t)$, which is then mapped back to a one-dimensional output signal $\mathbf{y}(t)$. Similar to RNNs, it has been found in previous work that Equation~\ref{SSM} also suffers from gradient vanish or exploding issues when modeling longer sequences. To tackle this issue, the work in~\cite{gu2021efficiently} leveraged HiPPO theory~\cite{gu2020hippo} to initialize the $\mathbf{A}$ matrix. HiPPO specifies a certain expression of $\mathbf{A}\in \mathbb{R}^{\textrm{N}\times \textrm{N}}$ (see Equation~\ref{HiPPO}), which allows the hidden state to memorize the input $\mathbf{u}(t)$ \footnote{Please refer to~\cite{gu2020hippo} for more details and relevant proofs.}. 

\begin{equation}
\label{HiPPO}
\text{HiPPO:}\ \mathbf{A}_{n,k}= -
\begin{cases}
  (2n+1)^{0.5}(2k+1)^{0.5} & \text{if}\ n>k \\
  n+1 & \text{if}\ n=k\\
  0  & \text{if}\ n<k,
\end{cases}
\end{equation}
where $n$ and $k$ indicate the row and column indices of $\mathbf{A}$. To implement Equation~\ref{SSM} using discrete inputs such as word or image tokens, the work in~\cite{gu2021efficiently} leverages the bi-linear discretization method~\cite{tustin1947method} and a discretized version of Equation~\ref{SSM} using a step size $\Delta$ is rewritten as:
\begin{align}
\begin{split}
\label{S4}
&\mathbf{x}_k = \Bar{\mathbf{A}}\mathbf{x}_{k-1}+\Bar{\mathbf{B}}\mathbf{u}_k \\
&\mathbf{y}_k=\Bar{\mathbf{C}}\mathbf{x}_k,
\end{split}
\end{align}
where $\Bar{\mathbf{A}}=(\mathbf{I}+\frac{\Delta\cdot \mathbf{A}}{2})/(\mathbf{I}-\frac{\Delta\cdot \mathbf{A}}{2})$, $\Bar{\mathbf{B}}=\Delta\cdot \mathbf{B}/(I-\frac{\Delta\cdot \mathbf{A}}{2})$ and $\Bar{\mathbf{C}} = \mathbf{C}$. Equation~\ref{S4} can be solved using a discrete convolution~\cite{gu2021efficiently}:
\begin{equation}
\label{S4_conv}
    \mathbf{y} = \Bar{\mathbf{K}} \circledast \mathbf{u},
\end{equation}
where $\mathbf{u}=\{u_0, u_1,\dots, u_{k-1}, u_k\}$ and $\Bar{\mathbf{K}} \in \mathbb{R}^\textrm{L} := \{\Bar{\mathbf{C}}\Bar{\mathbf{B}}, \Bar{\mathbf{C}}\Bar{\mathbf{A}}\Bar{\mathbf{B}},\dots,\Bar{\mathbf{C}}\Bar{\mathbf{A}}^{\textrm{L}-1}\Bar{\mathbf{B}}\}$ is a structured convolutional kernel and $\textrm{L}$ is the sequence length. Equation~\ref{S4_conv} is the core formulation of S$4$ model whose computational cost is linear to the input length and can be efficiently computed using fast Fourier transform (FFT) and inverse FFT. Moreover, to 
control the convolution kernel width, the work in~\cite{gu2021combining} set $\Delta$ as a learnable parameter.

\subsubsection{ViS4mer Model}

By utilizing the S4 model, the ViS4mer~\cite{islam2022long} achieves promising results in the long-form video understanding tasks. We start with defining some notations to help summarize the adaptation of S4 model in computer vision. Given a video clip $\mathbf{X} \in \mathbb{R}^{\textrm{H} \times \textrm{W} \times 3 \times \textrm{T}}$ consisting of $\textrm{T}$ RGB frames sampled from the video, we convert it into a sequence of $\textrm{S} \cdot \textrm{T}$ image tokens ${\mathbf{x}}_s^t \in  \mathbb{R}^\textrm{D}$ for $s=1, \hdots, \textrm{S}$ and $t=1, \hdots, \textrm{T}$. The tokens ${\mathbf{z}}_s^t$ are obtained by decomposing each frame into $\textrm{S}$ patches which are then projected to a $\textrm{D}$-dimensional space through a learnable linear transformation. This tokenization can be implemented by linearly mapping the RGB patches of each frame~\cite{bertasius2021space,neimark2021video}. Separate learnable positional encodings ${\mathbf{e}}_s$ and ${\mathbf{e}}^t$ are then applied to the patch embeddings ${\mathbf{z}}_s^t$ for the spatial and the temporal dimensions: ${\mathbf{x}}_s^t = {\mathbf{z}}_s^t + {\mathbf{e}}_s + {\mathbf{e}}^t$, formulating $\mathbf{x_{input}} = \{x_0^0, x_1^0, x_\text{S}^0, x_0^1,\dots,x_\text{S}^\text{T} \} $.

In ViS4mer~\cite{islam2022long}, a multi-scale S4 decoder is introduced for learning the long-term temporal reasoning. As is mentioned in $\S$~\ref{sss:preliminaries_s4}, S4 model has a linear computation and memory dependency with respect to the input length, which has significantly lower computational cost than the self-attention in transformers. The formulation of S4 decoder can be written as:
\begin{align}
\label{vis4mer}
\begin{split}
&{\bf x}_{s_4} = \text{S}_\text{4} \left(\text{LN}\left({\bf x}_{input}\right)\right) \\
&{\bf x}_{mlp} = \text{MLP} \left(\text{P}\left({\bf x}_{s_4}\right) \right) \\ 
&{\bf x}_{skip} = \text{Linear} \left(\text{P}\left({\bf x}_{input}\right) \right) \\
&{\bf x}_{out} = {\bf x}_{skip} + {\bf x}_{mlp},
\end{split}
\end{align}
Where $\text{LN}(\cdot), \text{MLP}(\cdot), \text{Linear}(\cdot)$ and $\text{P}(\cdot)$ represent the layer normalization~\cite{ba2016layer}, the multi-layer perception, linear layer and pooling layer, and ${\bf x}_{s_4}$ is the $\mathbf{y}$ in Equation~\ref{S4_conv}. 

\begin{figure}[]
\centering
\begin{subfigure}{.48\textwidth}
  \includegraphics[width=1.0\linewidth]{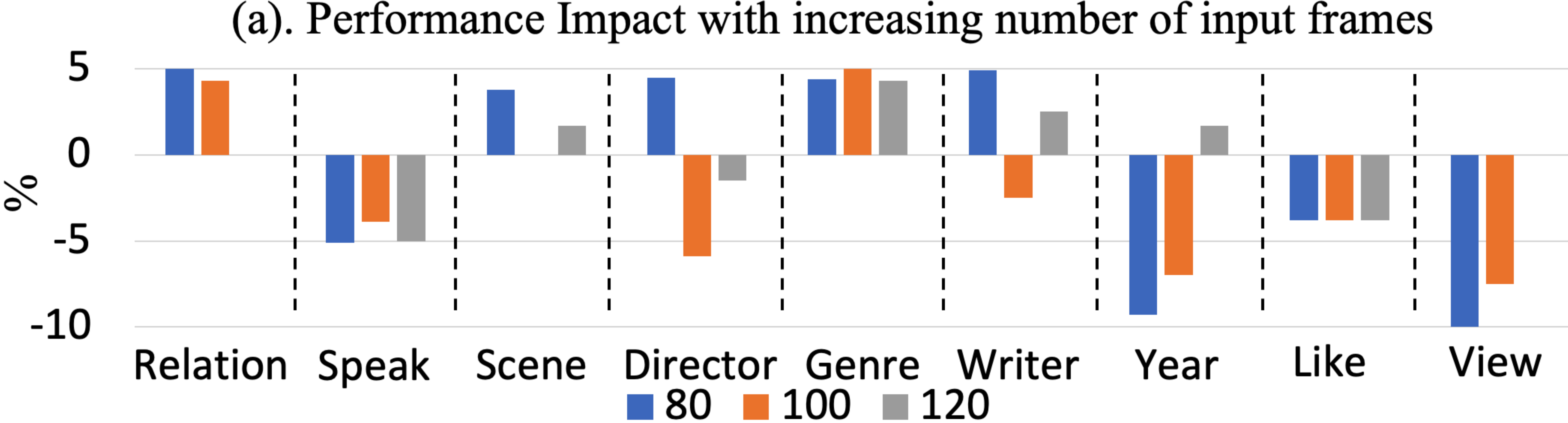}
\end{subfigure}
\begin{subfigure}{.48\textwidth}
  \includegraphics[width=1.0\linewidth]{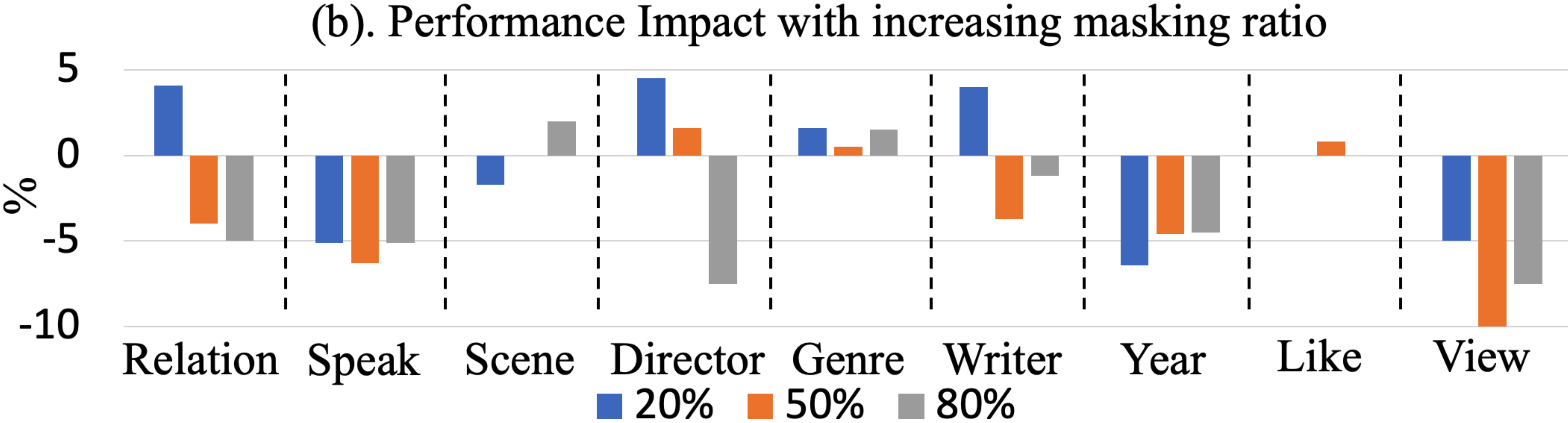}
\end{subfigure}
\caption{Performance gain/loss of ViS4mer on LVU dataset~\cite{lvu2021} with different settings of input frames and random masking ratio, where we conclude: \textbf{(a).} The performance is not substantially improved with increasing number of input frames. \textbf{(b).} Random masking strategy cannot effectively reduce redundant tokens.}
\label{concept_result}
\vspace{-2mm}
\end{figure}

\begin{figure*}[]
\centering
  \includegraphics[width=0.9\linewidth]{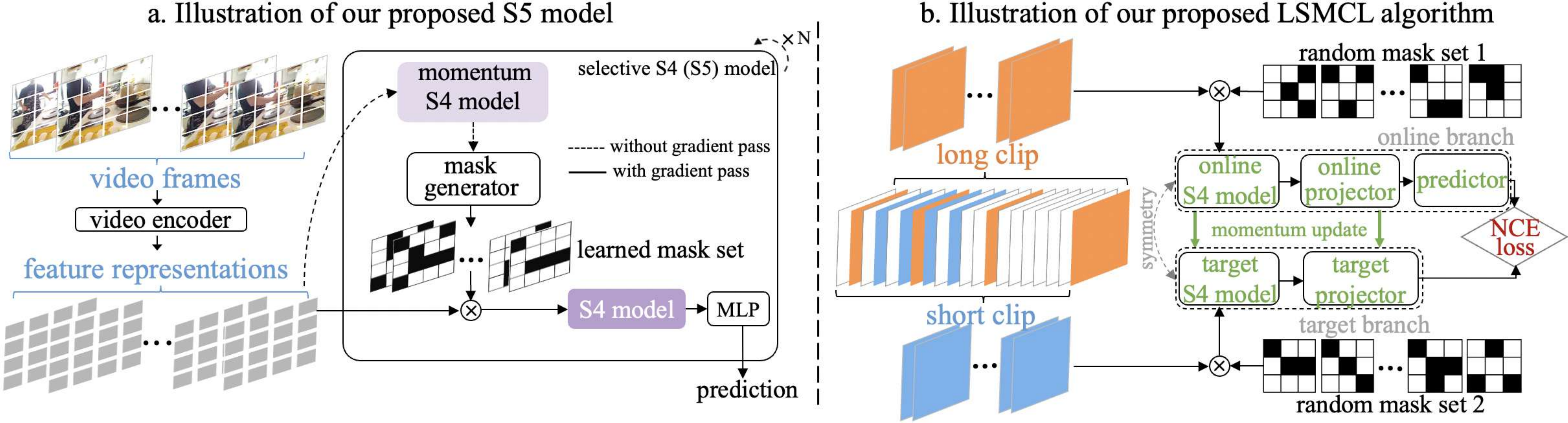}
\caption{(a) A visualization of our proposed S5 model. Compared to the S4 model, we introduce a selective token picking strategy ``mask generator'', leverageing the S4 feature from the momentum S4 model. The momentum S4 model is updated by the S4 model in the moving average manner. Both S4 model and momentum S4 model are consisted of a S4 layer~\cite{islam2022long, gu2021efficiently} and a LN layer~\cite{ba2016layer}. (b) An illustration of the proposed LSMCL pretraining framework, that initializes our S5 model to enrich the robustness.} 
\label{Fig:pipeline}
\vspace{-3mm}
\end{figure*}

\subsection{S4 Model in Long-form Video Understanding}
\label{ss:vis4mer_limitations}

\noindent To better understand the S4 model and long-form video understanding tasks, we re-implement ViS4mer~\cite{islam2022long} with different settings on LVU dataset~\cite{lvu2021} and demonstrate the result in Figure~\ref{concept_result}. From the observation that short-form video understanding tasks often benefit from longer input clips~\cite{bertasius2021space,fan2021multiscale,liu2021video,wang2022long}, we wonder if the performance of S4 model on different long-form video tasks would also be substantially improved with the increasing number of input frames. In Figure~\ref{concept_result} (a), we gradually increase the temporal extent from $60$ seconds to $120$ seconds. Compared to the performance of using 60 second input, we report the impact ratio of using $80$, $100$, $120$ second inputs in each task. From this Figure, we realize that not all long-form video tasks benefit from longer input context, and for those improved tasks, the performance is not necessarily improved with the longer input content. As a result, we raise the hypothesis that capturing long-term relationships is task- and data-dependent, and that additional performance improvements for those temporally-intensive tasks would also be hindered by the redundant spatiotemporal tokens produced by longer input content. Recalling Equation~\ref{S4} and~\ref{S4_conv}, each output token from S4 model is the result of structured discrete convolution for all previous inputs. Thus, we argue that treating all input token equally as ViS4mer~\cite{islam2022long} does is not appealing for S4 model to capture effective long-term dependencies, as not all tokens have the temporal relations and each task may also favor tokens in different space-time locations. To naively reduce the redundant tokens, we generate random masks on the 60 second input clips to drop tokens and increase the masking ratio from $20\%$ to $80\%$. Compared to the performance of un-masked input, we report the impact ratio of using random mask with masking ratio of $20\%$, $50\%$ and $80\%$ in Figure~\ref{concept_result} (b). Despite the minor improvement in some tasks, random masking degenerates the performance of most tasks, so it is not an effective method for reducing the redundancies. To this end, we are motivated to propose a selective S4 model which adaptively pick discriminative image tokens for the S4 model in different long-form video understanding tasks.

\subsection{Adaptive Token in Long-form Videos} 
\label{ss:adaptive_tokens}
\noindent To pick out discriminative image tokens from the long-form videos among various tasks, we extend the concept of adaptive token learning, formulating our Selective S5 (\textit{i.e.}, selective S4) model. Unlike previous image-based adaptive token learning works~\cite{yin2021adavit, meng2022adavit,rao2021dynamicvit,liang2022not} that rely on dense self-attention for capturing token-wise relationships, our S5 model avoids the self-attention computation in long-form videos by leveraging S4 features generated from the simulated linear time-invariant (LTI) system. Inherited from the linear complexity of the S4 model, our S5 model can receive long-form video token dependencies with low cost, thus making the adaptive token learning possible in long-form videos. In addition, we propose a momentum updated S4 model to dynamically produce S4 features from the long-form video data in different tasks. Figure~\ref{Fig:pipeline} (a) demonstrates the pipeline of our S5 model, where the momentum updated S4 model is the moving average of the S4 model.

Specifically, we cast our selective module in the S5 model as an adaptive mask learning problem. Given a mask generator $\text{MG}(\cdot)$ and its input ${\bf x}_{s_4}$, the mask generator is a lightweight architecture, which will be ablated in the Section~\ref{exp}. It will be trained for a classification task on predefined category space $\mathbb{C}=\{C_1,\dots,C_{\text{ST}}\}$, where $\text{S}\cdot \text{T}$ is the total number of image tokens in the video. Let's denote $p(c|{\bf x}_{s_4}) \in [0,1]$ be the normalized probabilistic output of $MG({\bf x}_{s_4})$, so that $\sum_{c=C_1}^{c=C_{\text{ST}}} p(c|{\bf x}_{s_4}) = 1$. Then, we sample $K$ categories without replacement from the probabilistic outputs of the mask generator. Finally, the $k^{th}$ selected image tokens can be written as:
\begin{align}
\begin{split}
\label{E_masking}
x_{\text{in}}^k = \mathbf{X^T}c^k
\end{split}
\end{align}
Where $\mathbf{X} \in \mathbb{R}^{\text{ST}\times D}$ represents $\text{S}\cdot \text{T}$ D-dimensional image tokens and $c^k$ is a one-hot vector that select $k^{th}$ token from the $\mathbf{X}$. The sampling process is important as it prevents the bias in the training that is potentially caused by the top-K selection. To make this sampling differentiable, we adopt the Gumbel-Softmax with Straight-Through tricks~\cite{jang2016categorical}, which is widely used in~\cite{lin2021vx2text,meng2022adavit}. Specifically, we introduce an additional gumbel noise $g \in \mathbb{R}^{1\times \text{ST}}$ into the predicted probability distribution $p \in \mathbb{R}^{1\times \text{ST}}$, where $g = -\log(-\log(u+\epsilon)+\epsilon)$ ($u\sim \text{Uniform(0,1)}$ , and $\epsilon$ is a small value for arithmetic robustness consideration). Then, we sample the top-K tokens from the re-parameterized distribution $p+g$. During the back-propagation, we estimate the gradient for each selected token $c$ as:
\begin{align}
\begin{split}
\label{straight-through}
 G\approx \bigtriangledown_{\text{MG}}\frac{\exp((\log p(c|{\bf x}_{s_4}) + g(c))/\rho)}{\sum_{c'=C_1}^{c'=C_{\text{ST}}}\exp((\log p(c'|{\bf x}_{s_4}) + g(c'))/\rho)}
\end{split}
\end{align}
where $\rho$ is the temperature factor controlling the sharpness.

\subsection{Long-Short Mask Contrastive Learning} 
\label{ss:LSMCL}
\noindent Previous token reduction/adaptive learning works rarely take model robustness into consideration. Informative tokens might be incorrectly dropped during training, which could hurt the performance of the model. In this paper, in addition to our proposed S5 model that explicitly picks informative tokens for various long-form video understanding tasks, we also propose Long-Short Mask Contrastive Learning (LSMCL) pretraining, which implicitly learns long-form video representations with better generalizability. Specifically, we equip the recent video contrastive learning framework LSTCL~\cite{wang2022long} with a random masking strategy on both long and short input clips, which mimics all possible scenarios that the selective module could produce in the S5 model. As a result, our S5 model with LSMCL pretraining would be more robust to and tolerant of errors from the selective module. Moreover, the long-short contrastive set-up will further improve the temporal predictability of our S5 model.

Formally, we sample a long clip $(x_L)$ and a short clip $(x_S)$ from each video sequence with largely different sampling strides $\tau_L$ and $\tau_S$, where $\tau_S < \tau_L$. Unlike LSTCL~\cite{wang2022long} and BraVe~\cite{recasens2021broaden} that apply independent random sampling, in our paper the temporal span of long clips includes the one of short clips, which prevents dissimilar semantics from two clips in long-form videos. Then, we independently generate binary random masks with a masking ratio of $\eta$ for each clip, which can be written as: $\mathcal{R}_\text{mask}(x,\eta), x\in \{x_L,x_S\}$. We set S4 model as the backbone of the query encoder $(f_q)$ and also adopt a momentum key encoder $(f_k)$ in the pipeline, which is widely accepted in MoCo~\cite{he2020momentum}, BYOL~\cite{grill2020bootstrap} and LSTCL~\cite{wang2022long}. Our query encoder and key encoder follow the same design with\cite{he2020momentum,grill2020bootstrap,wang2022long}, that consist of the backbone, projection and prediction heads. Denoting the parameter of $f_q$ as $\theta_q$ and the one of $f_k$ as $\theta_k$, we have: $\theta_k = m\theta_k + (1-m)\theta_q$, where $m \in [0,1]$ is a momentum coefficient. Similarly, the LSMCL adoptes similar objective as the InfoNCE~\cite{oord2018representation}:
\begin{align}
\begin{split}
&\text{Given:}\ q = f_q(\mathcal{R}_\text{mask}(x_S,\eta)), k = f_k(\mathcal{R}_\text{mask}(x_L,\eta))\\
&\mathcal{L}_{\text{LSMCL}} = \sum_{i} -\log\frac{\exp({q^i}^\top k^i/\rho)}{\exp({q^i}^\top k^i/\rho)+\sum_{j \neq i}\exp({q^i}^\top k^j/\rho)}\\
\end{split}
\end{align}
where $\rho$ is the temperature hyperparameter. As is commonly done in~\cite{chen2021empirical, chen2020exploring, grill2020bootstrap,caron2020unsupervised}, we symmetrize the loss function by switching $x_S$ and $x_L$ in $f_q$ and $f_k$. In our LSMCL, the S4 model is learned to find the correct step size $\Delta$ and SSM parameters to match the representation of random masked long and short clips. Given our S5 model takes adaptively learned image tokens in the downstream task, we believe the LSMCL could improve the robustness as well as the temporal modeling ability of S5 model when dealing with partially sampled image tokens. In Section~\ref{exp}, our S5 model with LSMCL empirically shows significantly improved results in long-form video understanding.
\begin{table*}[!htb]
\small
\centering
\begin{tabular}{|l|lll|llll|ll|l}
\cline{1-10} 
\multirow{2}{*}{\begin{tabular}[c]{@{}l@{}}Mask \\ Generator\end{tabular}} & \multicolumn{3}{c|}{Content ($\uparrow$)}                                       & \multicolumn{4}{c|}{Metadata ($\uparrow$)}                                                                    & \multicolumn{2}{c|}{User ($\downarrow$)}        \\ \cline{2-10} 
                                & \multicolumn{1}{l|}{Relation} & \multicolumn{1}{l|}{Speak} & Scene & \multicolumn{1}{l|}{Director} & \multicolumn{1}{l|}{Genre} & \multicolumn{1}{l|}{Writer} & Year  & \multicolumn{1}{l|}{Like} & View \\ \cline{1-10} 
No Mask (ViS4mer~\cite{islam2022long})                          & \multicolumn{1}{l|}{$57.14$}    & \multicolumn{1}{l|}{$40.79$} & $67.44$ & \multicolumn{1}{l|}{$62.61$}    & \multicolumn{1}{l|}{$54.71$} & \multicolumn{1}{l|}{$48.80$}  & $44.75$ & \multicolumn{1}{l|}{$0.26$} & $3.63$ \\ \cline{1-10} \cline{1-10} 
Random                          & \multicolumn{1}{l|}{$54.81$}    & \multicolumn{1}{l|}{$38.22$} & 67.44 & \multicolumn{1}{l|}{$63.60$}    & \multicolumn{1}{l|}{$54.97$} & \multicolumn{1}{l|}{$47.00$}  & $42.70$ & \multicolumn{1}{l|}{$0.25$} & $4.00$ \\ \cline{1-10} \cline{1-10} 
Single TX             & \multicolumn{1}{l|}{$57.85$}    & \multicolumn{1}{l|}{$40.79$} & $68.66$ & \multicolumn{1}{l|}{$63.98$}    & \multicolumn{1}{l|}{$55.12$} & \multicolumn{1}{l|}{$48.85$}  & $43.46$ & \multicolumn{1}{l|}{$0.26$} & $3.82$ &\tikzmark{e}  \\ 
$\textrm{Single TX}_{\textrm{S4}}$             & \multicolumn{1}{l|}{$\bf 60.54$}    & \multicolumn{1}{l|}{$\bf41.21$} & $\bf69.83$ & \multicolumn{1}{l|}{$\bf66.43$}    & \multicolumn{1}{l|}{$\bf57.55$} & \multicolumn{1}{l|}{$\bf49.47$}  & $\bf44.15$ & \multicolumn{1}{l|}{$\bf0.25$} & $\bf3.51$ &\tikzmark{f}$ \bf{\textcolor{ForestGreen}{+3.4}}$ \\ \cline{1-10} \cline{1-10} 

Stacked TXs         & \multicolumn{1}{l|}{$59.51$}    & \multicolumn{1}{l|}{$41.21$} & $69.83$ & \multicolumn{1}{l|}{$64.91$}    & \multicolumn{1}{l|}{$55.12$} & \multicolumn{1}{l|}{$51.83$}  & $47.55$ & \multicolumn{1}{l|}{$0.25$} & $3.42$ &\tikzmark{c} \\ 
$\textrm{Stacked TXs}_{\textrm{S4}}$          & \multicolumn{1}{l|}{$\bf61.98$}    & \multicolumn{1}{l|}{$\bf41.75$} & $\bf70.94$ & \multicolumn{1}{l|}{$\bf67.34$}    & \multicolumn{1}{l|}{$\bf59.16$} & \multicolumn{1}{l|}{$\bf51.83$}  & $\bf47.55$ & \multicolumn{1}{l|}{$\bf0.24$} & $\bf3.42$ &\tikzmark{d}$\bf\textcolor{ForestGreen}{+2.5}$ \\ \cline{1-10} \cline{1-10} 

Linear                          & \multicolumn{1}{l|}{$54.81$}    & \multicolumn{1}{l|}{$40.28$} & $67.44$ & \multicolumn{1}{l|}{$63.90$}    & \multicolumn{1}{l|}{$54.97$} & \multicolumn{1}{l|}{$48.17$}  & $42.77$ & \multicolumn{1}{l|}{$0.26$} & $3.95$ &\tikzmark{a} \\
$\textrm{Linear}_{\textrm{S4}}$                          & \multicolumn{1}{l|}{$\bf61.98$}    & \multicolumn{1}{l|}{$\bf41.75$} & $\bf69.88$ & \multicolumn{1}{l|}{$\bf66.40$}    & \multicolumn{1}{l|}{$\bf58.80$} & \multicolumn{1}{l|}{$\bf50.60$}  & $\bf47.70$ & \multicolumn{1}{l|}{$\bf0.25$} & $\bf3.51$ &\tikzmark{b}$\bf{\textcolor{ForestGreen}{+6.7}}$\\\cline{1-10} 
\end{tabular}
\begin{tikzpicture}[overlay, remember picture,  thick,  transform canvas={yshift=0.25\baselineskip,xshift=-0.25\baselineskip}]
    \draw [->] ({pic cs:a}) [bend left] to ({pic cs:b});
    \draw [->] ({pic cs:c}) [bend left] to ({pic cs:d});
    \draw [->] ({pic cs:e}) [bend left] to ({pic cs:f});
 \end{tikzpicture}
\caption{Performance of various mask generators in LVU~\cite{lvu2021} dataset, where we adopt 60 frames per clip and $50\%$ masking ratio. The bold results demonstrate the performance of using S4 feature ($x_{S_4}$ in Equation~\ref{vis4mer}). We also provide the average improvement ratio (in green) of nine jobs using S4 features compared to ViT features at the conclusion of each bold row. }
\label{MG_table}

\end{table*}

\section{Experiments}
\label{exp}
\subsection{Dataset}
\noindent\textbf{LVU dataset~\cite{lvu2021}:} is constructed from Movie Clip dataset~\cite{moiveclip}. It contains $\sim30K$ videos from $\sim3K$ movies. Each video lasts one to three minutes. The benchmark contains nine tasks covering a wide range of long-form video understanding tasks, which are further folded into three main categories: (i) content understanding, consisting of (‘relationship’, ‘speaking style’, ‘scene/place’) prediction, (ii) metadata prediction, including (‘director’, ‘genre’, ‘writer’, and ‘movie release year’) classification, and (iii) user engagement, predicting (‘YouTube like ratio’, and ‘YouTube popularity’). For classification and regression tasks, we report accuracy (for content understanding and metadata prediction) and mean-squared error (MSE) (for user engagement) as the evaluation metrics. 

\vspace{+2mm}
\noindent\textbf{COIN~\cite{tang2019coin, tang2020comprehensive}:} consists of 11,827 videos with 180 distinct procedural tasks, which are all collected from YouTube. These videos cover 12 domains, such as nursing $\&$ caring, vehicles, leisure $\&$ performance, gadgets, electric appliances, household items, science $\&$ craft, plants $\&$ fruits, snacks $\&$ drinks dishes, sports, and housework. The average length of a video is 2.36 minutes.

\vspace{+2mm}
\noindent\textbf{Breakfast~\cite{kuehne2014language}:} contains 1,712 videos of 10 complex cooking activities, which are performed by 52 different individuals in 18 different kitchens, resulting in over 77 hours of video footage. The averaged length of video in this dataset is around 2.7 minutes. Ten cooking activities include: making coffee, chocolate milk, juice, tea, cereals, fried egg, pancakes, fruit salad, sandwich and scrambled egg.

\subsection{Implementation Details}
\noindent Following~\cite{islam2022long, lvu2021}, we stack three structure blocks, which share similar structure to that described in Equation~\ref{vis4mer}, and sample video frames at 1 fps. Unlike previous work, we include an adaptive mask generator to effectively pick image tokens before feeding the input into S4 model. As the advantages of our S5 model will naturally be diminished on less redundant sequences, we follow the same architecture of ViS4mer~\cite{islam2022long} but adopt the S5 model as the first block. 
For data argumentation, we resize each video frame to the spatial resolution of $224\times224$ and use a patch size of $16\times16$. In addition, we use ViT-L~\cite{dosovitskiy2020image} pretrained on ImageNet-21K~\cite{krizhevsky2012imagenet} as the feature extractor in the LVU dataset; Swin-B~\cite{liu2021swin} pretrained on Kinetics-600~\cite{kay2017kinetics} as the feature extractor in COIN and Breakfast datasets. The size of the input in each dataset is also the same as~\cite{islam2022long}: we adopt 60-second input for the LVU dataset and 64-second input for the COIN and Breakfast datasets. In the LSMCL, we adopt the setting from LSTCL~\cite{wang2022long} and apply independent global random masking on long and short clips, which share the same masking ratio with the adaptive mask generator. Unless otherwise noted, we conduct our ablation studies on the LVU dataset due to its diverse tasks in the long-form video understanding. Finally, we report the best performance of our model on all three datasets and compare with the previous state-of-the-art works.
\subsection{Ablation Study}
\noindent\textbf{a. Our S5 is better than S4 and random masking:} To demonstrate the effectiveness of our proposed S5 model, we compare the performance of S4 models with no mask, random mask, and mask generators of different architectures. Specifically, we utilize one Transformer (TX), two stacked Transformers (TXs), and one linear layer as the mask generator and evaluate on 9 tasks on the LVU dataset (Table~\ref{MG_table}). In addition, we also evaluate the effectiveness of using S4 features from the momentum-updated S4 model. For each architecture, we compare the result of using ViT features and S4 features as the mask generator input. As can be seen from the Table~\ref{MG_table}, the performance of each task substantially increases with the computational complexity of the mask generator. 
Results show our design significantly outperforms ViS4mer~\cite{islam2022long} and the random masking strategy, and the performance of each task is further improved by using S4 features. Notably, the mask generator with one linear layer achieves on par performance to one of the more complex transformer architectures. 

\begin{figure}[]
\centering
  \includegraphics[width=0.7\linewidth]{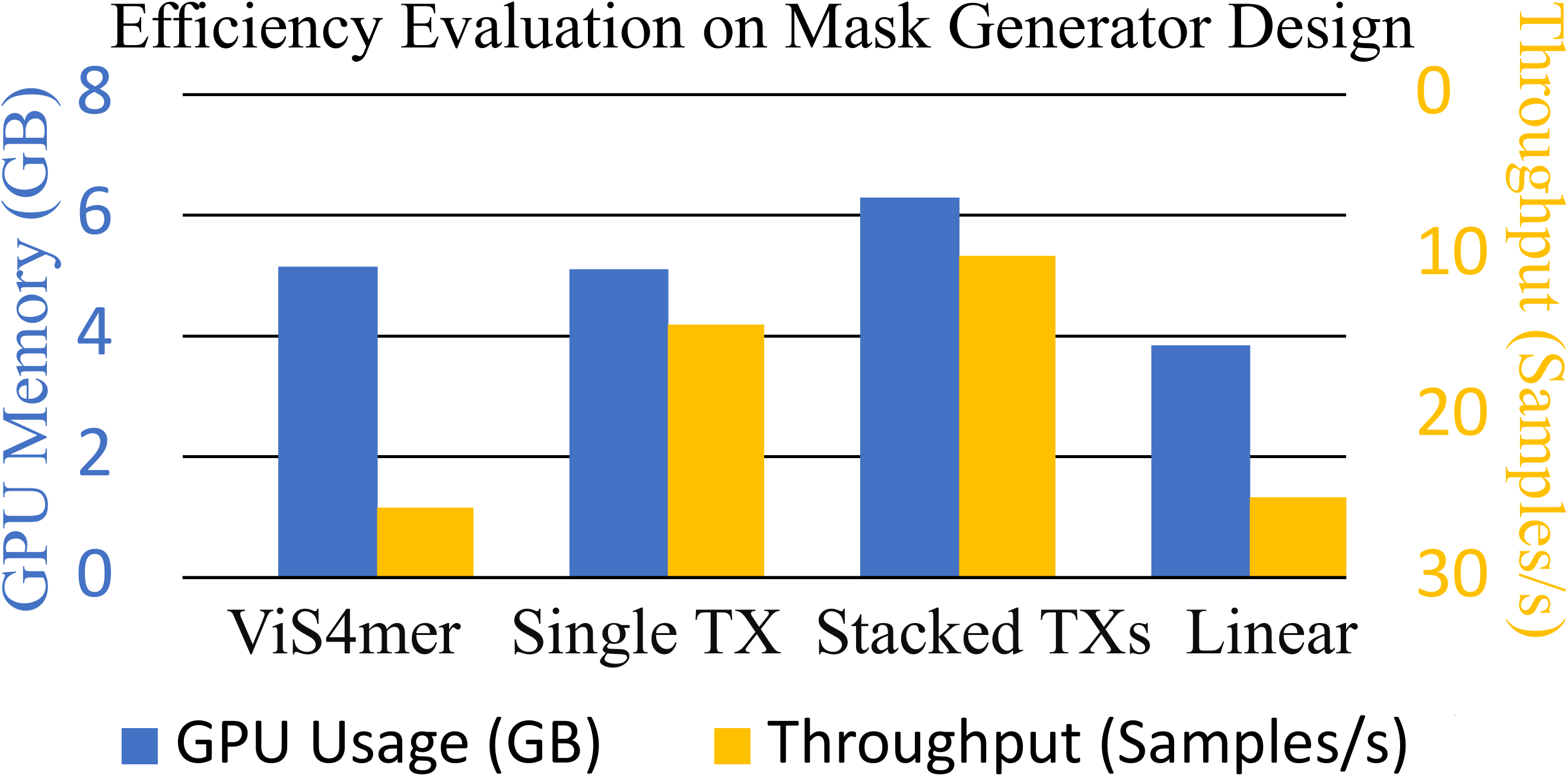}
\caption{Efficiency evaluation of each method in Table~\ref{MG_table}, which demonstrates the GPU memory usage as well as throughput. Our proposed S5 model with linear mask generator saves $25\%$ memory cost and achieves on par throughput with ViS4mer~\cite{islam2022long}.}
\label{efficiency}
\end{figure}

\vspace{+2mm}
\noindent\textbf{b. Our S5 reduces up to $25\%$ memory usage:} In Figure~\ref{efficiency}, we also demonstrate the efficiency of our S5 model with the different masking architectures mentioned previously. Compared to ViS4mer (the one without masking strategies) using same number of input frames, our S5 model with linear mask generator reduces the memory footprint by $25\%$ while maintaining the same level of throughput. Memory consumption and throughput are not improved by the intricate transformer mask generators. Since the linear mask generator has a smaller memory footprint and performs tasks more effectively overall, we use it in our S5 model in the following experiments.

\begin{figure*}[!htb]
\centering
\begin{subfigure}{.245\textwidth}
  \includegraphics[width=0.98\linewidth]{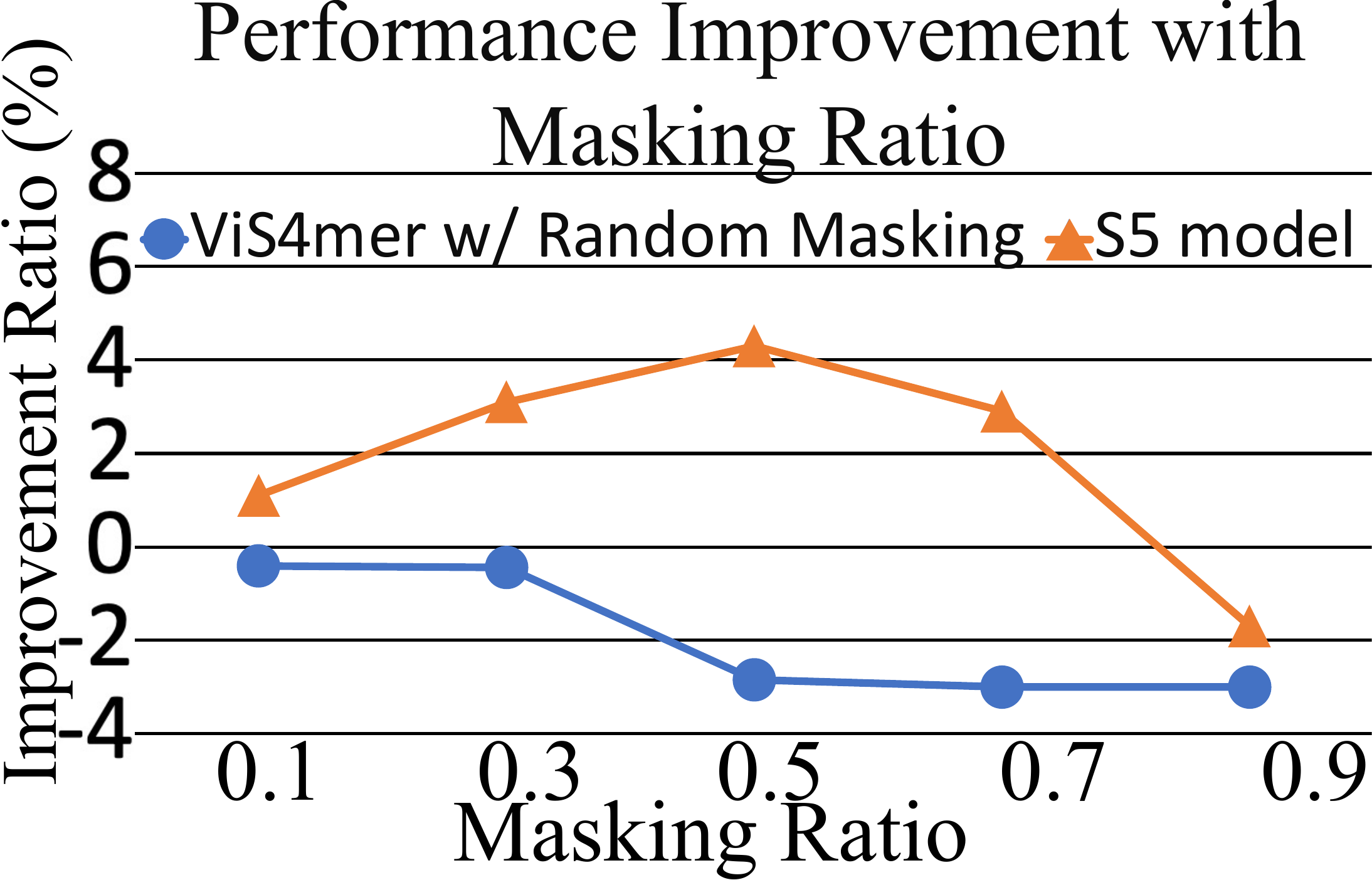}
  \caption{}
  \label{MG_masking}
\end{subfigure}
\begin{subfigure}{.245\textwidth}
  \includegraphics[width=0.98\linewidth]{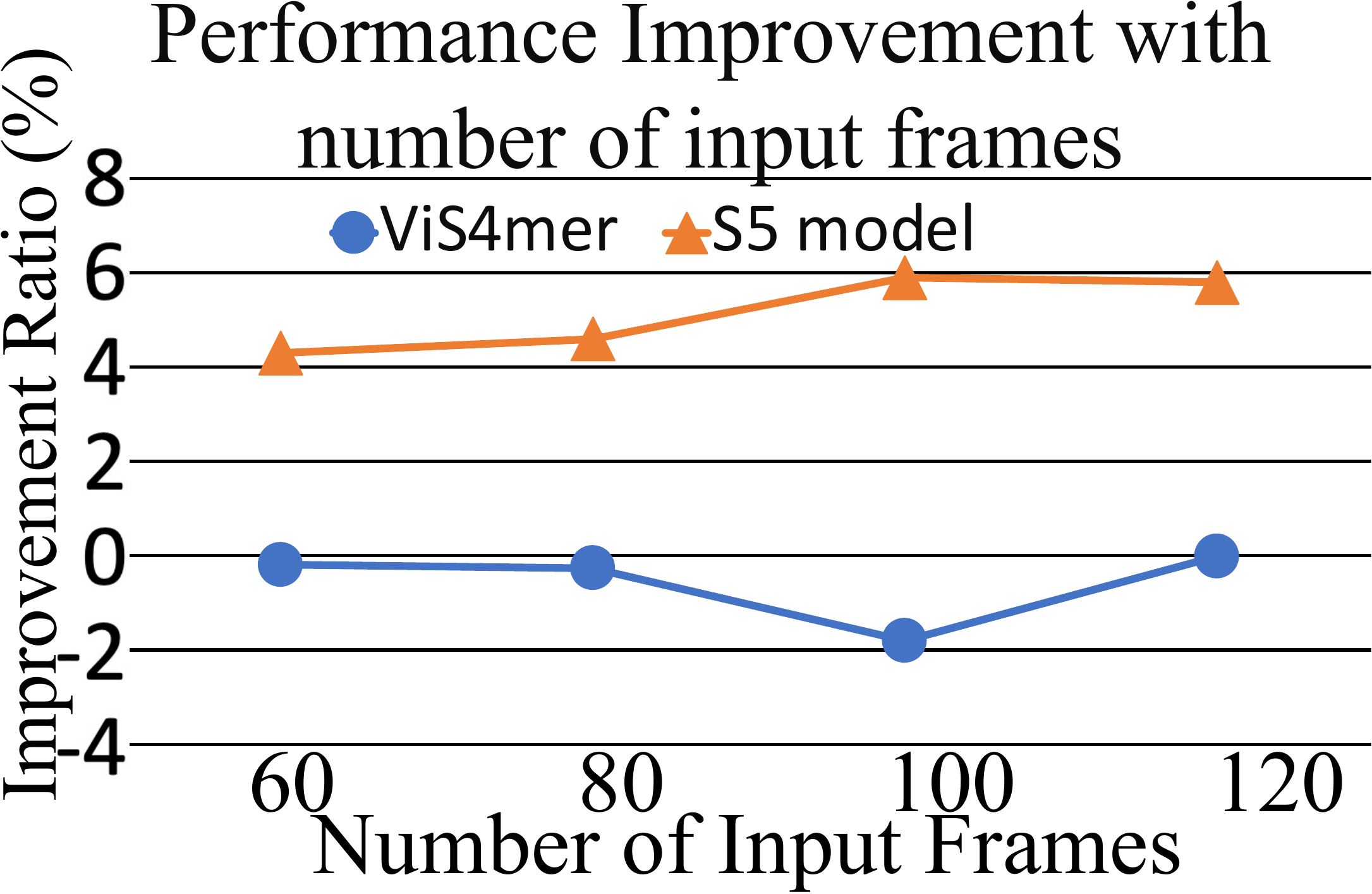}
    \caption{}
  \label{MG_seq_len}
\end{subfigure}
\begin{subfigure}{.245\textwidth}
  \includegraphics[width=0.95\linewidth]{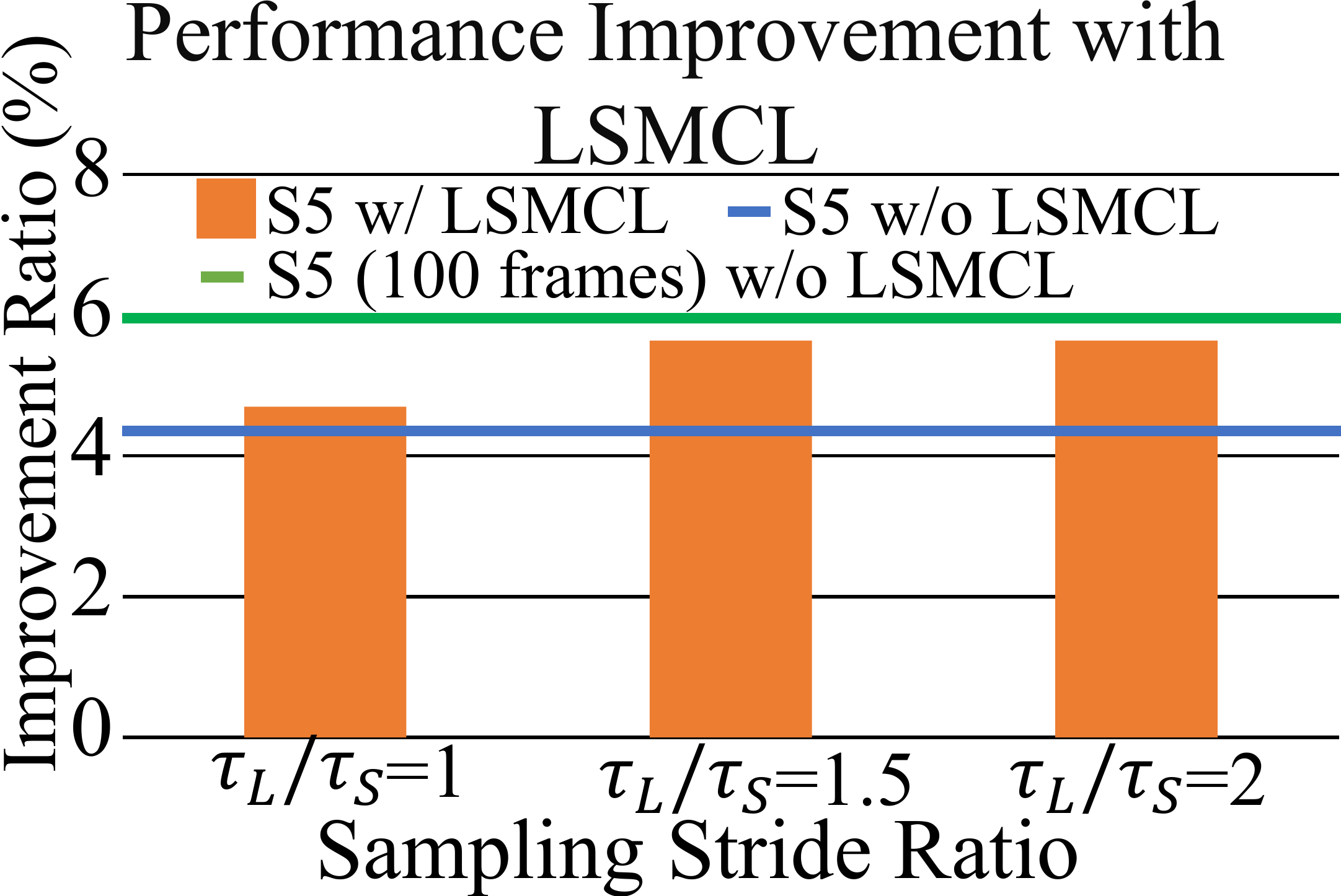}
    \caption{}
  \label{LSMCL_stride}
\end{subfigure}
\begin{subfigure}{.245\textwidth}
  \includegraphics[width=0.98\linewidth]{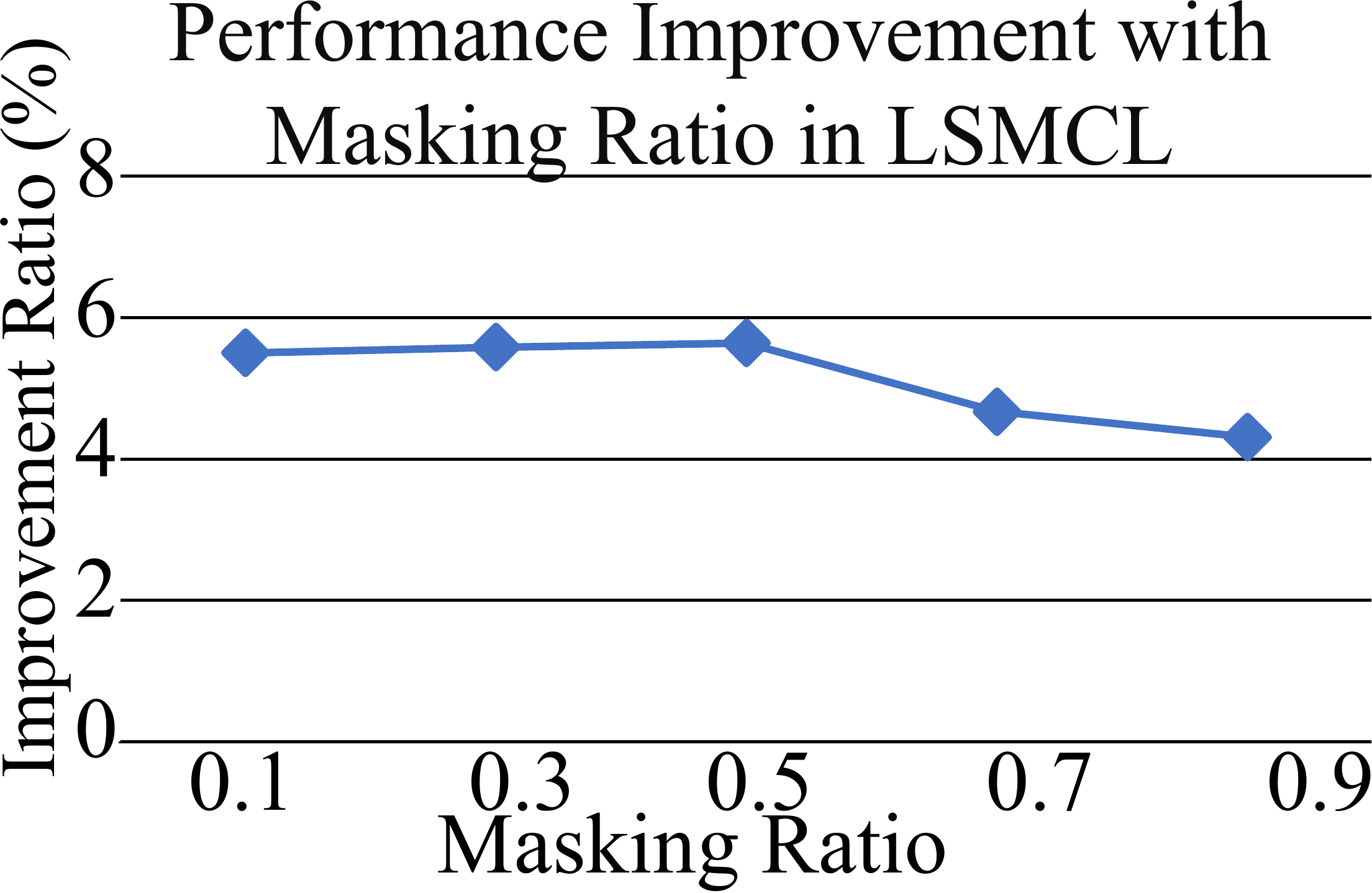}
    \caption{}
  \label{LSMCL_masking}
\end{subfigure}
\caption{Compared to the baseline performance, average improvement performance of our method on LVU dataset. Unless otherwise noted, the default number of input frame and masking ratio is 60 and $50\%$. \textbf{(a).} We compared our S5 model and S4 model with random masking with increasing masking ratio; \textbf{(b).} We compare our S5 model and S4 model with increasing number of input frames; \textbf{(c).} We show the effect of LSMCL pretraining with different long-short sampling stride ratio. In addition, we provide the performance of S5 model without LSMCL and S5 model with 100 input frames; \textbf{(d).} We show the impact of the increasing masking ratio in the LSMCL pretraining.}
\label{allinone}
\end{figure*}

\vspace{+2mm}
\noindent\textbf{c. Impact of Masking Ratio and Sequence Length:} In Figure~\ref{MG_masking} and~\ref{MG_seq_len}, we study the effect of masking ratio and sequence length with our S5 model. We set ViS4mer~\cite{islam2022long} (60 frames without mask generator) as baseline and report the average improvement percentage of 9 tasks on LVU dataset by using S5 model with variant masking ratio/sequence length. To demonstrate the effectiveness of our S5 model, we also compare the performance of ViS4mer~\cite{islam2022long} with different settings in these two figures. Figure~\ref{MG_masking} clearly shows that the performance of our S5 model increases initially as the masking ratio increases, which indicates that our selective model effectively picks informative image tokens for the S4 model. However, the performance starts to drop dramatically when the masking ratio is over $50\%$. This is because when the masking ratio increases to be above certain level, the informative tokens are forced to be dropped. As a result, we adopt $50\%$ masking ratio in our following experiments. In Figure~\ref{MG_seq_len}, we observe substantial improvement of S5 model with increasing number of input frames. In contrast to the performance of ViS4mer~\cite{islam2022long}, our proposed S5 model is indeed able to capture longer term dependencies while reducing the spatial-temporal redundancy in the input. 

\vspace{+2mm}
\noindent\textbf{d. Effect of Multiple S5 models:} As shown in Figure~\ref{Fig:pipeline}, multiple S5 models can be stacked in the pipeline, similar to what is commonly done in Transformer~\cite{dosovitskiy2020image, bertasius2021space, wu2022memvit} and ViS4mer~\cite{islam2022long}. In the previous setup, we only adopt one S5 model, leaving the remaining blocks as S4 models. By stacking multiple S5 models, we find a further $\bf 0.5\%$ average improvement on the LVU dataset. Less redundant sequences will inevitably reduce the performance gain from our S5 model, decreasing the benefit from stacking additional S5 blocks. As a result, we utilize only one S5 model after the video encoder for maximum memory efficiency gain and throughput.

\vspace{+2mm}
\noindent\textbf{e. Ablation on LSMCL:} In Figure~\ref{LSMCL_stride} and~\ref{LSMCL_masking}, we evaluate the effectiveness of our proposed LSMCL with different sampling strides and random masking ratios. For both figures, we set the performance of ViS4mer~\cite{islam2022long} as the baseline and report the average improvement ratio (in percentage) of 9 tasks from LVU with different settings. From Figure~\ref{LSMCL_stride}, our S5 model with LSMCL can achieve better performance even when $\tau_{L}=\tau_{S}$, which suggests that LSMCL can increase the robustness of our S5 model and help it handle incorrectly picked tokens. When we gradually increase the $\frac{\tau_{L}}{\tau_{S}}$, the performance of S5 model is further improved as the model is able to capture longer temporal context via the proposed LSMCL. Indeed, the performace using LSMCL approaches the performance without LSMCL \textbf{with $\textbf{66\%}$ more input frames} (shown in Figure~\ref{MG_seq_len} both around $6\%$ boost). In Figure~\ref{LSMCL_masking}, we further ablate the random masking ratio used in LSMCL. When the masking ratio of LSMCL is over $50\%$, the benefit from LSMCL is insignificant as the input does not provide sufficient information. Thus, we consider $50\%$ masking ratio in LSMCL for better efficiency in the long-form video contrastive learning.


\begin{table*}[!bpht]
\small
\centering
\begin{tabular}{|l|lll|llll|ll|l|}
\hline
\multirow{2}{*}{Model}   & \multicolumn{3}{c|}{Content ($\uparrow$)}                                                               & \multicolumn{4}{c|}{Metadata ($\uparrow$)}                                                                                                    & \multicolumn{2}{c|}{User ($\downarrow$)}                          & \multicolumn{1}{c|}{\multirow{2}{*}{\begin{tabular}[c]{@{}c@{}}GPU Usage\\(GB) ($\downarrow$)\end{tabular}}} \\ \cline{2-10}
                                                       & \multicolumn{1}{l|}{Relation}       & \multicolumn{1}{l|}{Speak}          & Scene          & \multicolumn{1}{l|}{Director}       & \multicolumn{1}{l|}{Genre}          & \multicolumn{1}{l|}{Writer}         & Year           & \multicolumn{1}{l|}{Like}          & View          & \multicolumn{1}{c|}{}                                                                               \\ \hline
Obj. T4mer~\cite{lvu2021}                                                         & \multicolumn{1}{l|}{54.76}          & \multicolumn{1}{l|}{33.17}          & 52.94          & \multicolumn{1}{l|}{47.66}          & \multicolumn{1}{l|}{52.74}          & \multicolumn{1}{l|}{36.30}          & 37.76          & \multicolumn{1}{l|}{0.30}          & 3.68          & N/A                                                                                               \\ \hline
Performer~\cite{choromanski2020rethinking}                                                                   & \multicolumn{1}{l|}{50.00}          & \multicolumn{1}{l|}{38.80}          & 60.46          & \multicolumn{1}{l|}{58.87}          & \multicolumn{1}{l|}{49.45}          & \multicolumn{1}{l|}{48.21}          & 41.25         & \multicolumn{1}{l|}{0.31}          & 3.93         & 5.93                                                                                              \\ \hline
Orthoformer~\cite{patrick2021keeping}                                                                   & \multicolumn{1}{l|}{50.00}          & \multicolumn{1}{l|}{38.30}          & 66.27        & \multicolumn{1}{l|}{55.14}          & \multicolumn{1}{l|}{55.79}          & \multicolumn{1}{l|}{47.02}          & 43.35         & \multicolumn{1}{l|}{0.29}          & 3.86         & 5.56                                                                                                \\ \hline
VideoBERT~\cite{sun2019videobert}                                                                   & \multicolumn{1}{l|}{52.80}          & \multicolumn{1}{l|}{37.90}          & 54.90          & \multicolumn{1}{l|}{47.30}          & \multicolumn{1}{l|}{51.90}          & \multicolumn{1}{l|}{38.50}          & 36.10          & \multicolumn{1}{l|}{0.32}          & 4.46          & N/A                                                                                                 \\ \hline
LST~\cite{islam2022long}                                                                               & \multicolumn{1}{l|}{52.38}          & \multicolumn{1}{l|}{37.31}          & 62.79          & \multicolumn{1}{l|}{56.07}          & \multicolumn{1}{l|}{52.70}          & \multicolumn{1}{l|}{42.26}          & 39.16          & \multicolumn{1}{l|}{0.31}          & 3.83          & 41.38                                                                                                \\ \hline
ViS4mer~\cite{islam2022long}                                                                           & \multicolumn{1}{l|}{57.14}          & \multicolumn{1}{l|}{40.79}          & 67.44          & \multicolumn{1}{l|}{62.61}          & \multicolumn{1}{l|}{54.71}          & \multicolumn{1}{l|}{48.80}          & 44.75          & \multicolumn{1}{l|}{0.26}          & 3.63          & 5.15                                                                                                 \\ \hline\hline
$\textrm{Ours}_{\textrm{60 frames}}$                                                                          & \multicolumn{1}{l|}{\textbf{61.98}}          & \multicolumn{1}{l|}{\textbf{41.75}} & \textbf{69.88} & \multicolumn{1}{l|}{\textbf{66.40}} & \multicolumn{1}{l|}{\textbf{58.80}} & \multicolumn{1}{l|}{\textbf{50.60}} & \textbf{47.70} & \multicolumn{1}{l|}{\textbf{0.25}} & \textbf{3.51} & \textbf{3.85}                                                                          \\ \hline      
$\textrm{Ours}_{\textrm{60 frames+LSMCL}}$                                                                           & \multicolumn{1}{l|}{\textbf{61.98}}          & \multicolumn{1}{l|}{\textbf{41.75}} & \textbf{72.53} & \multicolumn{1}{l|}{\textbf{66.40}} & \multicolumn{1}{l|}{\textbf{61.34}} & \multicolumn{1}{l|}{\textbf{50.60}} & \textbf{47.70} & \multicolumn{1}{l|}{\textbf{0.24}} & \textbf{3.51} & \textbf{3.85}                                                                           \\ \hline 
$\textrm{Ours}_{\textrm{100 frames}}$                                                                            & \multicolumn{1}{l|}{\textbf{66.71}}          & \multicolumn{1}{l|}{\textbf{41.78}} & \textbf{73.28} & \multicolumn{1}{l|}{\textbf{66.64}} & \multicolumn{1}{l|}{\textbf{63.65}} & \multicolumn{1}{l|}{\textbf{50.60}} & \textbf{47.85} & \multicolumn{1}{l|}{\textbf{0.25}} & \textbf{3.51} & \textbf{3.95}                                                                                \\ \hline 
$\textrm{Ours}_{\textrm{100 frames+LSMCL}}$                                                                             & \multicolumn{1}{l|}{\textbf{67.11}}          & \multicolumn{1}{l|}{\textbf{42.12}} & \textbf{73.49} & \multicolumn{1}{l|}{\textbf{67.32}} & \multicolumn{1}{l|}{\textbf{65.41}} & \multicolumn{1}{l|}{\textbf{51.27}} & \textbf{47.95} & \multicolumn{1}{l|}{\textbf{0.24}} & \textbf{3.51} & \textbf{3.95}                                                                                \\ \hline 
\end{tabular}

\caption{Comparison to the state-of-the-art methods on LVU dataset testing set.}
\label{sota_LVU}
\end{table*}

\begin{table}[]
\small
\centering
\begin{tabular}{|l|l|l|l|}
\hline
Method              & P.T. Dataset & P.T. Samples & Accuracy \\ \hline
TSN~\cite{tang2020comprehensive}                 & Kinetics-400        & 306K                & 73.40    \\ \hline
D-Sprv.~\cite{lin2022learning}       & HowTo100M           & 136M                & 90.00    \\ \hline
ViS4mer~\cite{islam2022long}             & Kinetics-600        & 495K                & 88.41    \\ \hline\hline
Ours                & Kinetics-600        & 495K                & \textbf{90.42}    \\ \hline
$\textrm{Ours}_{\textrm{+LSMCL}}$                & Kinetics-600        & 495K                & \textbf{90.81}    \\ \hline
\end{tabular}
\caption{Comparison to the state-of-the-art methods on COIN dataset. P.T. stands for pretraining.}
\label{sota_COIN}
\vspace{-4mm}
\end{table}

\begin{table}[]
\small
\centering
\begin{tabular}{|l|l|l|l|}
\hline
Method              & P.T. Dataset & P.T. Samples & Accuracy \\ \hline
VideoGraph~\cite{hussein2019videograph}                 & Kinetics-400        & 306K                & 69.50    \\ \hline
Timeception~\cite{hussein2019timeception}                 & Kinetics-400        & 306K                & 71.30   \\ \hline
GHRM~\cite{zhou2021graph}                 & Kinetics-400        & 306K                & 75.50   \\ \hline
D-Sprv.~\cite{lin2022learning}       & HowTo100M           & 136M                & 89.90    \\ \hline
ViS4mer~\cite{islam2022long}             & Kinetics-600        & 495K                & $\text{85.10}^{*}$    \\ \hline\hline
Ours                & Kinetics-600        & 495K                & \textbf{90.14}    \\ \hline
$\textrm{Ours}_{\textrm{+LSMCL}}$               & Kinetics-600        & 495K                & \textbf{90.70}    \\ \hline
\end{tabular}
\caption{Comparison to the state-of-the-art methods on Breakfast dataset. P.T. stands for pretraining. $^{*}$We were not able to reproduce the $88.17\%$ baseline result reported in~\cite{islam2022long}, but our proposed S5 model still largely improves from $85.10\%$, and achieves the new state-of-the-art result.}%
\label{sota_Breakfast}
\vspace{-4mm}
\end{table}

\subsection{Comparison with the State-Of-The-Arts}
\noindent In Table~\ref{sota_LVU}, we compare our method on LVU dataset with previous state-of-the-art methods. Specifically, the LST~\cite{islam2022long} adopt the same architecture with ours, but substitutes the S5/S4 model to the transformer architecture. Whereas the Performer~\cite{choromanski2020rethinking} and Orthoformer~\cite{patrick2021keeping} apply the efficient attention in the transformer architecture, that do not require quadratic complexity \textit{w.r.t.} the input length. When compared to baseline ViS4mer~\cite{islam2022long}, we achieve up to $\mathbf{9.6\%}$ improvement. When compared to other methods, ours outperforms by an even more significant margin. This shows that our method is consistently more effective in understanding the long-form videos.

To demonstrate the generalizability of our method, we evaluate our S5 model on COIN~\cite{tang2019coin,tang2020comprehensive} and Breakfast~\cite{kuehne2014language} datasets, which are challenging long-range procedural activity classification datasets. Our proposed method achieves $\textbf{2.4\%}$ and $\textbf{5.5\%}$ over the ViS4mer~\cite{islam2022long} and outperforms the other state-of-the-arts by $\textbf{0.81\%}$ and $\textbf{0.80\%}$ respectively. Notice that D-Sprv.~\cite{lin2022learning} leverages HowTo100M dataset~\cite{miech2019howto100m} for pretraining, which volume is much larger than our pre-training dataset (Kinetics-600~\cite{carreira2018short}). Putting together the aforementioned performance gain and memory efficiency gain, our S5 model successfully demonstrates its efficiency and effectiveness in learning discriminative representation via selecting informative image tokens from long-form video sequences.

\section{Conclusion}
\noindent In this paper, we proposed a selective structured state-space sequence (S5) model for long-form video understanding, where we adopt a lightweight mask generator to adaptively pick informative tokens from long-form videos. Our mask generator avoids dense self-attention computation as what is applied in previous works. It leverages the sequential output of the simulated linear time invariant (LTI) system, and benefits from the momentum distillation of S4 model, enabling our S5 model to dynamically learn from informative tokens for different long-form video tasks. To mitigate the negative impact of picking less informative tokens, we also propose a LSMCL pretraining to improve the robustness and further broaden the temporal horizon of our model. Through extensive experiments, we demonstrate the effectiveness of each proposed component in our S5 model, achieving the new state-of-the-art performance in three challenging long-form video understanding benchmarks.


{\small
\bibliographystyle{ieee_fullname}
\bibliography{egbib}
}
\newpage 
\appendix
\section{Implementation Details}
\noindent In addition to the implementation details introduced in Section 4.2 of the main paper, we provide more information of training our S5 model and LSMCL below.
\subsection{S5 model}
\noindent Following ViS4mer~\cite{islam2022long}, we introduce a MLP layer in each block to reduce the feature dimension by a factor of $2\times$. Each MLP layer is consisted of a linear layer, a GELU activation layer~\cite{hendrycks2016gaussian} and a dropout layer, where the dropout rate is 0.2. For updating the momentum S4 model, we explore different values of momentum coefficient and set it as $0.01$ to produce the best performance. For all of our experiments of S5 model, we use AdamW optimizer~\cite{loshchilov2017decoupled} with a learning rate of $10^{-3} \times \frac{\text{batch size}}{16}$, and with a weight decay of $0.01$. For COIN~\cite{tang2019coin, tang2020comprehensive}, Breakfast~\cite{kuehne2014language} and each task on LVU dataset~\cite{lvu2021}, we train our S5 model for 100 epochs and reduce the learning rate by a factor of $0.2$ when the training loss has stopped reducing in the past $1$ epoch. We train our S5 model by using $8\times$ NVIDIA Tesla V100 $16\text{G}$ GPUs with a batch size of $16$. All the implementations are coded with PyTorch~\cite{paszke2019pytorch}.

\subsection{LSMCL}
\noindent In LSMCL, we sample two clips with different sampling strides from the video sequence, and the clip shape is consistent with the one in finetuning the S5 model. Specifically, we sample input clips of size $60\times 3 \times 224 \times 224$ on LVU dataset~\cite{lvu2021} and $64\times 3 \times 224 \times 224$ on COIN~\cite{tang2019coin, tang2020comprehensive} and Breakfast datasets~\cite{kuehne2014language}. The sampling stride ratio is set to be $\frac{\tau_L}{\tau_S}=1.5$, which is also ablated in the Figure 5(c) in the main paper. Following LSTCL~\cite{wang2022long}, we adopt a query encoder and a key encoder in the LSMCL. The query encoder consists of a S4 model backbone, a MLP projection head and an additional prediction MLP head. The purpose of the prediction layer is to transform the representation of the query clip to match the key. The key encoder consists of a S4 model backbone and a MLP projection head. The momentum coefficient for updating the key encoder is $0.99$. Following~\cite{wang2022long,chen2021empirical}, the MLP projection head has $3$ layers while the MLP prediction head has $2$ layers. The hidden layers of both MLPs are $4096$-D and are with ReLU; the output layers of both MLPs are $256$-D, without ReLU. In LSMCL, all layers in both MLPs have BN~\cite{ioffe2017batch}, which follows~\cite{chen2021empirical,chen2020simple}. In terms of the optimizer, we adopt AdamW~\cite{loshchilov2017decoupled} with a learning rate of $10^{-4} \times \frac{\text{batch size}}{256}$, and with a weight decay of $0.05$. We train our LSMCL for $300$ epochs in total, and adopt learning rate warm-up~\cite{goyal2017accurate} for the first $40$ epochs. We train LSMCL by using $8\times$ NVIDIA Tesla V100 $16\text{G}$ GPUs with a batch size of $64$ and we optimize the model with the loss in Equation 8, where the $\rho=0.2$. 

\begin{figure}[!htb]
\centering
  \includegraphics[width=0.65\linewidth]{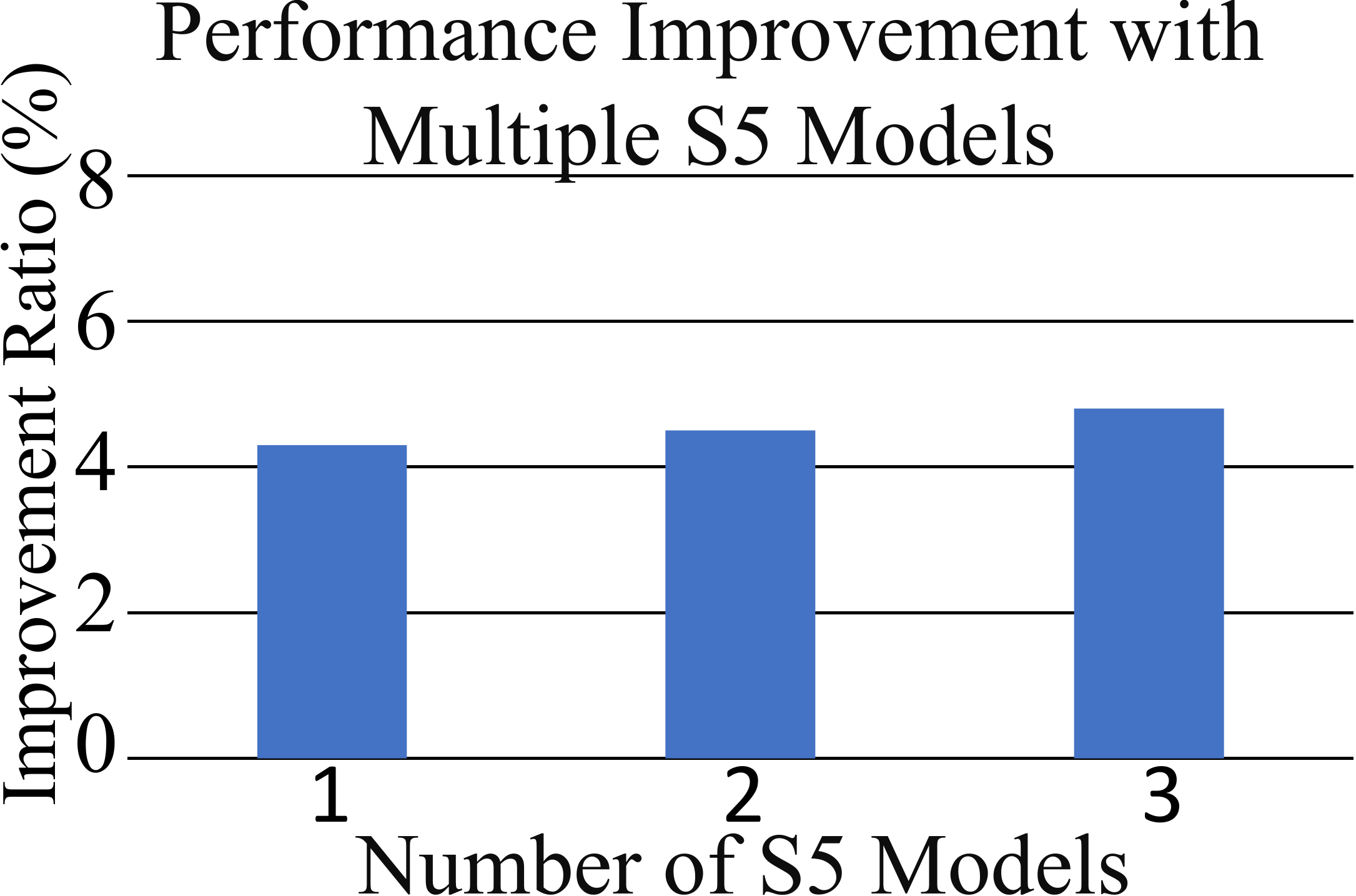}
\caption{Compared to the baseline performance, average improvement performance of our method with different settings on LVU dataset. Unless otherwise noted, the default number of input frame and masking ratio is 60 and $50\%$. We study the effect of leveraging multiple S5 models in our work, where we substitutes more S4 model in orignial ViS4mer~\cite{islam2022long} with our S5 model.}
\label{multiS5}
\end{figure}


\section{Effectof Multiple S5 Models}
\noindent In this paper, we improve the previous S4 model by introducing a novel selective module, formulating the Selective S4 (S5) model. For fair comparison, we follow the architecture introduced in the ViS4mer~\cite{islam2022long}, which utilizes three S4 models with pooling and MLP layers in between. As the advantages of our S5 model will naturally be diminished on less redundant sequences, our default setting is to substitute the first S4 model in ViS4mer~\cite{islam2022long} with our proposed S5 model while keep the rest architecture the same with ViS4mer~\cite{islam2022long}. In this section, we study the impact of using more S5 models in the ViS4mer~\cite{islam2022long} architecture. In Figure~\ref{multiS5}, we gradually increase the number of blocks that use S5 model instead of S4 model. We set the performance of ViS4mer as the baseline, and report the averaged improvement percentage over 9 tasks on LVU dataset~\cite{lvu2021}. Compared to the method of using S4 models, our method achieves substantial improvement by including more S5 models. However, less duplicated sequences will definitely result in a decrease in our S5 model's performance gain, which will lessen the advantage of stacking additional S5 blocks. 
\end{document}